\documentclass[10pt,twocolumn,letterpaper]{article}

\usepackage{wacv}              

\usepackage{graphicx}
\usepackage{amsmath}
\usepackage{amssymb}
\usepackage{booktabs}

\usepackage[
    backend=biber,
    style=ieee,
  ]{biblatex}
\usepackage{pifont}

\usepackage[breaklinks,colorlinks]{hyperref}

\usepackage[capitalize]{cleveref}
\crefname{section}{Sec.}{Secs.}
\Crefname{section}{Section}{Sections}
\Crefname{table}{Table}{Tables}
\crefname{table}{Tab.}{Tabs.}

\def\wacvPaperID{2319} 
\def\confName{WACV}
\def\confYear{2024}

\begin{document}

\title{DeVOS: Flow-Guided Deformable Transformer for Video Object Segmentation}

\author{Volodymyr Fedynyak\textsuperscript{1}, Yaroslav Romanus\textsuperscript{1,2}, Bohdan Hlovatskyi\textsuperscript{1}, Bohdan Sydor\textsuperscript{1},\\
Oles Dobosevych\textsuperscript{1}, Igor Babin\textsuperscript{1,2}, Roman Riazantsev\textsuperscript{2}
\and
\textsuperscript{1}Ukrainian Catholic University, \textsuperscript{2}ADVA Soft
\and
{\tt\small \{v.fedynyak, yaroslav.romanus, bohdan.hlovatskyi, b.sydor, dobosevych\}@ucu.edu.ua}\\
{\tt\small \{ihor.babin, roman.riazantsev\}@adva-soft.com}\\
}

\maketitle

\begin{abstract}
   The recent works on Video Object Segmentation achieved remarkable results by matching dense semantic and instance-level features between the current and previous frames for long-time propagation. Nevertheless, global feature matching ignores scene motion context, failing to satisfy temporal consistency. Even though some methods introduce local matching branch to achieve smooth propagation, they fail to model complex appearance changes due to the constraints of the local window. In this paper, we present DeVOS (Deformable VOS), an architecture for Video Object Segmentation that combines memory-based matching with motion-guided propagation resulting in stable long-term modeling and strong temporal consistency. For short-term local propagation, we propose a novel attention mechanism ADVA (Adaptive Deformable Video Attention), allowing the adaption of similarity search region to query-specific semantic features, which ensures robust tracking of complex shape and scale changes. DeVOS employs an optical flow to obtain scene motion features which are further injected to deformable attention as strong priors to learnable offsets. Our method achieves top-rank performance on DAVIS 2017 val and test-dev (88.1\%, 83.0\%), YouTube-VOS 2019 val (86.6\%) while featuring consistent run-time speed and stable memory consumption.
\end{abstract}

\section{Introduction}
\label{sec:intro}

Video Object Segmentation (VOS) is a fundamental task of video understanding. In a semi-supervised approach, it is formulated as the identification and segmentation of objects through the video sequence given the ground truth annotation masks for the first and, optionally, some other frames.

Previous VOS methods \cite{oh2019stm, cheng2021stcn, cheng2021mivos, seong2021hierarchical, wang2022look} focus on distilling the information from past frames into a \textit{feature memory} storage and then perform a dense memory matching to identify objects on the current frame. Some approaches \cite{xie2021rmnet, yang2020CFBI, yang2020CFBIP} suggest enhancing memory-based matching with mask propagation to achieve smooth predictions and improve temporal consistency. \citeauthor*{yang2021associating} in their work \citetitle{yang2021associating} (AOT) \cite{yang2021associating} proposed using image attention mechanism \cite{vaswani2017attention} to perform hierarchical propagation and matching, employing global attention for memory readouts and local windowed attention for short-term propagation. In DeAOT \cite{yang2021decoupling}, the architecture was further improved by decoupling processing of visual and object information.

 Recently, \citeauthor{wang2022look} in ISVOS \cite{wang2022look} noticed that existing methods suffer from performance degradation in scenarios of substantial shape deformations and appearance changes caused by camera and scene motion. The authors propose to utilize instance discriminative features while performing dense matching with the memory bank, ensuring selecting of the correct object from past frames and avoiding false positives. ISVOS achieves state-of-the-art performance on most of the benchmarks, outperforming the methods specifically designed for long-time videos, \textit{e.g.}, XMem \cite{cheng2022xmem} and AFB-URR \cite{NEURIPS2020_liangVOS} on the Long-time Video dataset. However, the main research effort of the aforementioned approach lies in determining \textit{how to improve features for matching} without focusing on \textit{how exactly to perform the matching}.

 \begin{figure*}[ht]
    \centering
  \includegraphics[width=0.6\linewidth]{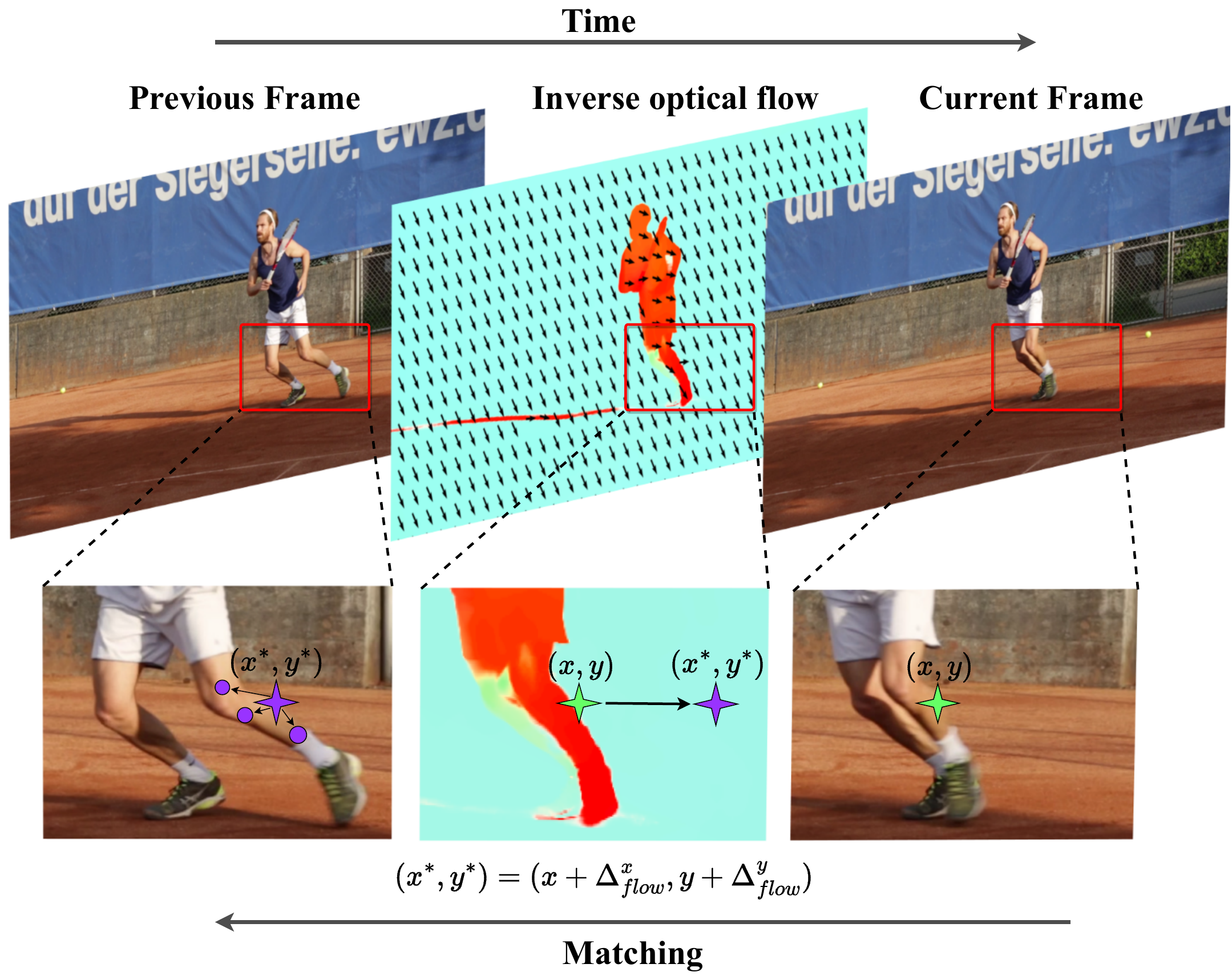}
  \caption{The process of matching features between the current and preceding frames is divided into two steps: flow-based displacement adjustment and semantics-driven deformable attention}
  \label{fig:devos_intro}
\end{figure*}
 
The temporal evolution of an object's appearance depends on the semantic properties, \textit{i.e.} rigidity. Thus it's natural to adapt the similarity search region to specific semantic features of the query point. Some existing implementations of matching logic construct a global affinity matrix between current and previous features and use similarity score as a matching objective. STCN \cite{cheng2021stcn} and ISVOS \cite{wang2022look} adopt negative L2 distance for this purpose, treating all possible search locations equally. XMem \cite{cheng2022xmem} proposes anisotropic L2 similarity, allowing query-specific importance interpretation. AOT \cite{yang2021associating} shrinks the search space by using windowed cross-attention for short-term matching, while the query-specific importance is assured by learnable relative position bias. In existing methods the query-specific adaptation is limited to only tweaking importance over the query-agnostic set of spatial locations
(often limited to a region around the spatial location of a query). In terms of handling motion, global matching leads to degenerating of temporal-spatial consistency, while windowed matching fails to capture rapid movement.

We argue that adapting similarity search region to specific query semantic properties is crucial to perform propagation robust to appearance, scale, and shape change. To further enhance the performance, we propose to decouple motion and semantics during matching, adopting a global scene displacement field as an initial offset of the search region.

In this spirit, we present DeVOS, a novel architecture for VOS introducing a new attention-based short-term matching mechanism ADVA (Adaptive Deformable Video Attention). Inspired by Deformable DETR \cite{zhu2020deformable} and DAT \cite{Xia_2022_CVPR}, we adopt multi-scale deformable cross-attention capable of sampling search locations on the previous frame based on motion and query-specific semantic features of the current frame. More specifically, given some reference location, we predict initial global offset using the scene motion features, \textit{positioning} the search region. Consequently, we use corresponding query features to predict several local offsets, \textit{shaping} the search region. Finally, keys and values are sampled from predicted locations using the previous frame and passed to multi-head attention. Comparing to previous methods, we present formulation of deformable cross-attention for video-related tasks, while preserving efficient query offset modelling. ADVA is described in details in \cref{sec:deform}. Furthermore, we enhance the keys and queries of the matched video frames with motion features to achieve strong temporal consistency, which is described in details in \cref{sec:qk}. The short-term and long-term memory matching results are fused and passed to the decoder producing the final propagated object mask. To obtain motion features a generic optical flow estimation network is used.

We conduct experiments on the standard DAVIS \cite{Perazzi2016} and YouTube-VOS \cite{xu2018youtubevos1} benchmarks. We optionally conduct additional training on the large-scale MOSE 2023 \cite{MOSE} dataset to achieve robustness under complex VOS scenarios. Conducted experiments demonstrate that DeVOS achieves top-ranked performance while enjoying consistent run-time speeds. It is worth noting that our research direction is orthogonal to those in ISVOS \cite{wang2022look}, DeAOT \cite{yang2021decoupling}, and XMem \cite{cheng2022xmem} and can further benefit from the ideas presented in those works.

\section{Related Work}
\label{sec:relworks}

\begin{figure*}[ht!]
    \centering
  \includegraphics[width=\linewidth]{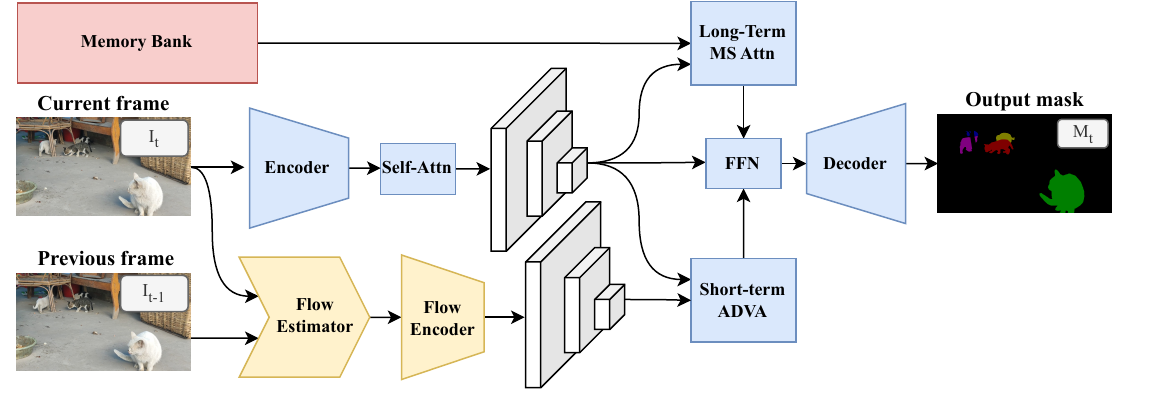}
  \caption{The overview of DeVOS architecture. The current frame is processed through encoder and self-attention block. After that, optical flow between current and previous frames is computed for the adaptive deformable video attention between current and previous frame features. Information from a memory bank containing frames for long-term memory is incorporated through a long-term multi-scale deformable attention block.}
  \label{fig:devos_arch}
\end{figure*}

\subsection{Optical Flow Estimation}

Optical flow estimation is crucial for modeling global motion. Initial studies focused on optimization problems, emphasizing visual similarity and regularization \cite{horn1981determining, black1993framework, bruhn2005lucas, sun2014quantitative}. The introduction of deep neural networks, especially convolutional networks, significantly advanced this field.

The RAFT model \cite{teed2020raft} introduced a significant upgrade to optical flow estimation, incorporating the multi-scale search window through the recurrent module. Following the introduction of RAFT, subsequent studies like GMA \cite{jiang2021learning} and DEQ-Flow \cite{bai2022deep} further improved accuracy and computational efficiency. FlowFormer \cite{huang2022flowformer} extends RAFT by utilizing a transformer-based strategy for aggregating cost volume in latent space, building on Perceiver IO \cite{jaegle2021perceiver}. It pioneered the use of transformers \cite{vaswani2017attention} for long-range relationships in optical flow, achieving top-tier performance.

Recently, \citeauthor{Fedynyak_2023_CVPR} in WarpFormer \cite{Fedynyak_2023_CVPR} showed that employing an optical flow estimator to support a generic VOS architecture by warping the past frames into the current frame domain could be benefitial for smooth propagation.

\subsection{Video Object Segmentation}

A key approach in the field of Video Object Segmentation (VOS) is AOT (Associating Objects with Transformers for VOS) \cite{yang2021associating}. This method uses a Long Short-Term Transformer (LSTT) block that incorporates short-term attention and long-term attention mechanisms to extract features from input imagery. Long-term attention gathers information from extended memory frames, while short-term attention disseminates information from the previous frame. The outputs of both attention units are integrated into a feed-forward network and then passed to the decoder to predict the current object mask.

DeAOT \cite{yang2021decoupling} builds on the hierarchical propagation concept of AOT for semi-supervised video object segmentation, introducing a dual-branch propagation for object-agnostic and object-specific embeddings.

XMem \cite{cheng2022xmem} is a Video Object Segmentation (VOS) architecture designed for long videos. It utilizes the Atkinson-Shiffrin memory model to create an architecture with multiple independent, interconnected feature memory stores. It incorporates a sensory memory, a working memory, and a long-term memory. A memory potentiation algorithm is used to consolidate working memory elements into long-term memory, preventing memory overload and maintaining performance for long-term prediction.

\begin{figure*}[ht!]
    \centering
  \includegraphics[width=0.85\linewidth]{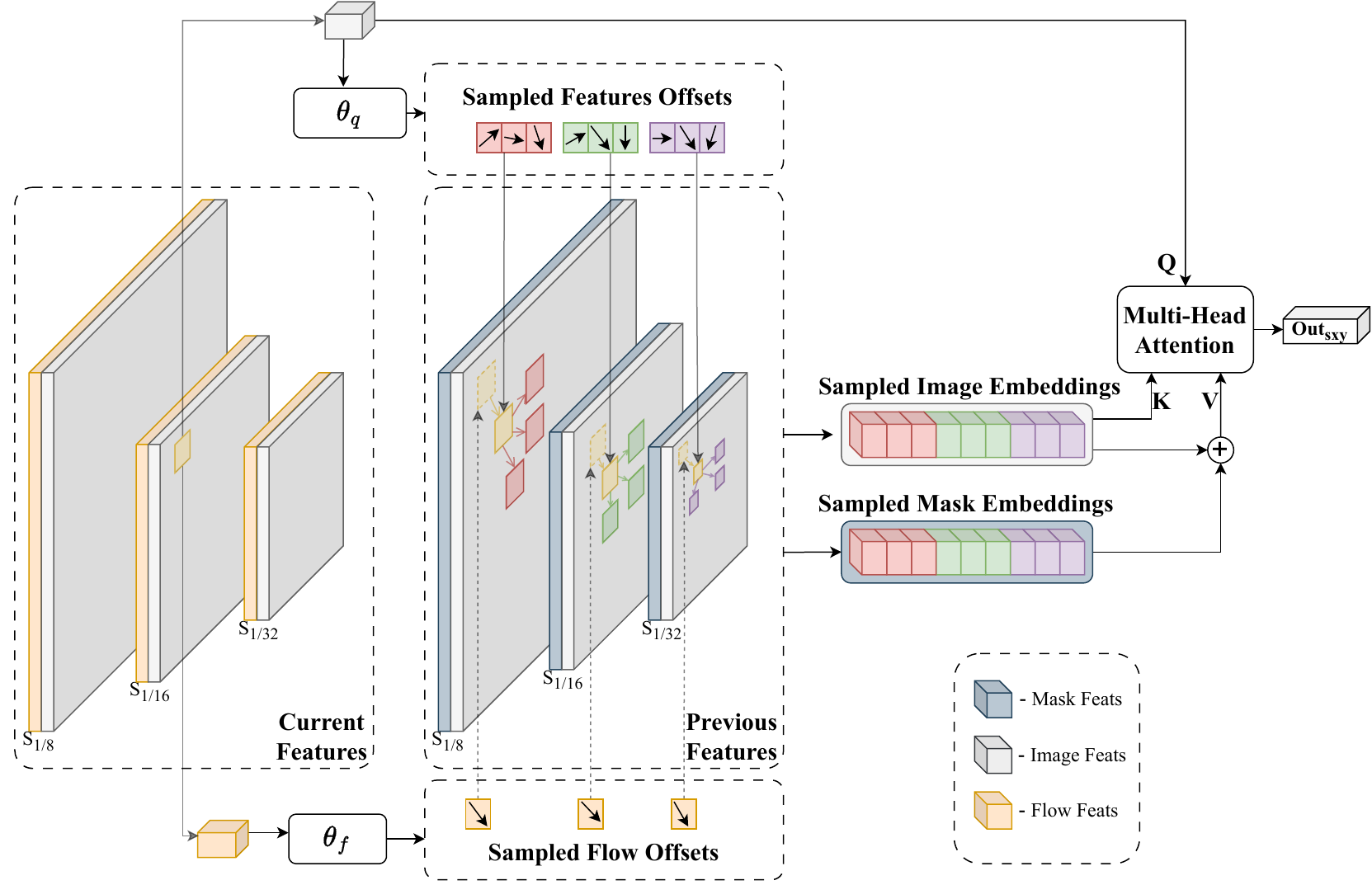}
  \caption{Adaptive deformable video attention. The multi-scale flow-based feature matching consists of two steps: offsets prediction for features alignment and multi-head attention. Two types of offsets are used: flow-based offsets for movement compensation and semantic-based offsets to extract previous frame image and mask embeddings. Multi-head attention combines the previous frame mask, and image embeddings based on the correlation of the previous frame sampled features and query image embedding vector.}
  \label{fig:devos_msattn}
\end{figure*}

The paper ISVOS \cite{wang2022look} further highlights the importance of instance understanding in VOS. While recent memory-based methods have achieved impressive results in VOS through dense matching between current and past frames, these methods often falter when confronted with large appearance variations or viewpoint changes caused by object and camera movements. To mitigate these issues, the authors propose a two-branch network for VOS, which incorporates a query-based instance segmentation (IS) branch to delve into the instance details of the current frame. This approach allows the integration of instance-specific information into the query key, facilitating instance-augmented matching. These works collectively underscore the importance of instance understanding in VOS and propose solutions that effectively integrate this concept into existing VOS methods.

\subsection{Vision Transformers and Deformable Attention}

Transformers have gained traction in computer vision, yet their large receptive fields pose computational and memory challenges. Deformable attention, introduced in Deformable DETR \cite{zhu2020deformable} and Deformable Attention Transformer (DAT) \cite{Xia_2022_CVPR}, addresses these issues by focusing on a small set of key sampling points, reducing computational load and enhancing performance.

Deformable DETR applies deformable attention in the detection head to improve performance on small objects and speed up convergence. DAT introduces a deformable self-attention module in the vision backbone, enabling data-dependent selection of key and value pairs, thus efficiently capturing more informative features and modeling long-range relations.

\section{Method}
\label{sec:method}

To describe our method, let's consider a video sequence denoted as $V = [X_1, X_2, ..., X_T]$, along with the annotation mask of the first frame. Our approach processes the frames sequentially, storing the predicted results in memory to inform future predictions. Firstly, we extract features from the current image, $X_t$, using a backbone encoder, resulting in a feature map $I_t$. Subsequently, the feature map of the current image is compared with the memory frames to perform semantic matching and propagate the mask. Finally, the matching result, combined with features from the encoder at multiple scales, is fed into the decoder, which restores the object mask in the original resolution (see the full architecture in Figure \ref{fig:devos_arch}).

\subsection{Adaptive Deformable Video Attention}
\label{sec:deform}

The classic attention operation is defined as follows:
\[\mbox{Att}(Q, K, V) = \mbox{Corr}(Q, K)V = \operatorname{softmax}\Bigl(\frac{QK^T}{\sqrt{C}}\Bigr)V\]
where $Q$, $K$, and $V$ denote the attention queries, keys, and values, respectively, while $C$ is the embedding space dimension. 
In order to choose only the relevant spatial locations, a data-driven approach is used for selecting some predefined number $N_k$ of key and value pairs. For each query and its corresponding reference point, $N_k$ offsets are learned to indicate the specific locations from which values and keys should be sampled. These offsets are obtained based on the query features, ensuring that they capture and represent semantic information:

\[\Delta p_{qk} = \theta_{\operatorname{offset}}(Q W^Q), \]

where $\theta_{\operatorname{offset}}$ - sub-network for offset generation. To stabilize the training, the predicted offsets are scaled to fit into window with size $\sigma$: $\Delta p_{qk} \leftarrow \sigma \cdot \mbox{tanh}(\Delta p_{qk})$. After offsets are sampled attention is computed as in classical formulation:

\[\mbox{DfAttn}(Q, K, V, p_q) = \operatorname{softmax}\Bigl(\frac{Q K_p^T}{\sqrt{C}}\Bigr)V_p ,\]
\[p \in \{p_q + \Delta p_{qk}: k \in K\}\]

where subscript $k$ refers to an index of learned offset for a given query. Such formulation allows learning sparse regions to attend to each of the queries and can be naturally extended to high-resolutions as it is linear with respect to spatial resolution. Following the \cite{zhu2020deformable} and \cite{vaswani2017attention}, we extend this formulation with multi-scale feature maps and multi-head attention correspondingly. For each resolution, we add both learnable positional embeddings $\pi$ and scale-level embeddings $\omega$. Moreover, the window size $\sigma$ is dynamically adjusted in proportion to the scale.

Motion becomes a crucial factor when designing deformable cross-attention for multiple frames. As the nature of the movement is isotropic, in the same naive formulation, query offsets would be forced to learn windowed attention. This is unwanted as it undermines offsets' ability to learn query-specific information and thus - similarity search region adaptation. To mitigate this issue, we propose to decouple \textit{motion} and \textit{semantic} information, creating separate offset branches for them:

\[\Delta p_{qk}^q = \theta_{\operatorname{offset}}^q(Q W^Q), \Delta p_{qk}^f = \theta_{\operatorname{offset}}^f(F_{inv} W^F),\]
where $F_{inv}$ denotes inverse optical flow and $\theta_{\operatorname{offset}}^q, \theta_{\operatorname{offset}}^f$ - sub-networks for offset generation based on queries and flow respectively. Besides, we normalize the predicted flow offsets to fit into the image: $\Delta p_{qk}^f \leftarrow D \cdot \mbox{tanh}(\Delta p_{qk})$, where $D$ denotes spatial dimension size. Afterward, the total offsets are computed as the combined sum of semantic(query-based) and motion(flow-based) offsets:

\[\Delta p_{qk} = \Delta p_{qk}^q + \Delta p_{qk}^f\]

This novel type of attention, called adaptive deformable video attention (ADVA), adapts search region for cross-attention based on the motion and semantic features, thus showing superior perfomance on VOS benchmarks. We believe that is can be applied to various video-related tasks beyond VOS as well.

\subsection{QK-flow}
\label{sec:qk}

To further leverage motion information, we explore the possibility of integrating it into the semantic feature map. We argue that this enhancement helps in distinguishing between different instances of the same semantic class, as they naturally have distinct motion patterns. Formally, we denote the direct and inverse flow between the previous and current frames as $F_{dir}$ and $F_{inv}$. To integrate motion information, we augment our queries ($Q$) with the linearly projected flow towards the previous frame: $Q_m = Q + W_{inv}F_{inv}$. Similarly, we augment our keys ($K$):  $K_m = K + W_{dir}F_{dir}$. Here, the subscript $m$ indicates that our queries and keys have been enriched with motion information.

\subsection{Multi-scale matching}
To benefit from the sparsity of the proposed attention formulation to its fullest, we propose to conduct semantic matching with memory bank on multi-scale feature maps. We argue, that it helps dealing with overlapping objects that share a similar appearance thanks to effective utilization of high-resolution features. Formally, our backbone encoder generates features at multiple scales, denoted as $[I_t^{(1)}, I_t^{(2)}, I_t^{(3)}]$, corresponding to scales of $\frac{1}{8}$, $\frac{1}{16}$, and $\frac{1}{32}$, respectively. To ensure consistent matching, we map the features on different scales into the same embedding space with linear projections.

Subsequently, our multi-scale features are passed to a self-attention block, implemented as deformable self-attention \cite{zhu2020deformable}. For short-term matching, we employ sparse attention in the form of the ADVA, which is described in \cref{sec:deform}. During the long-term matching, we stack the flattened encoder feature maps on the spatial dimensions of $\frac{1}{16}$ and $\frac{1}{32}$, then perform an attention-based global matching with the memory bank.

\subsection{Network details}

\begin{table*}[ht!]
\caption{The quantitative evaluation on multi-object benchmarks YouTube-VOS 2019 and DAVIS 2017. \textsuperscript{\textasteriskcentered} denotes training on MOSE 2023. \textsuperscript{\textdagger} denotes replacing ResNet50 with Swin-B encoder. \textsuperscript{\textdaggerdbl} denotes FPS retimed on our hardware. Top-3 results are denoted in bold font.}
\label{tab:main}
\vspace{10pt}
\centering
  \begin{tabular}{l|ccccc|ccc|ccc|c}
    \toprule[2pt]
    & \multicolumn{5}{c|}{YouTube-VOS 2019 Val} & \multicolumn{3}{c|}{DAVIS 2017 Val} & \multicolumn{3}{c|}{DAVIS 2017 Test}  \\
    \midrule
    Methods & $\mathcal{J}_s$ & $\mathcal{F}_s$ & $\mathcal{J}_u$ & $\mathcal{F}_u$ & Avg & $\mathcal{J}$ & $\mathcal{F}$ & Avg & $\mathcal{J}$ & $\mathcal{F}$ & Avg & FPS \\
    \midrule[1.5pt]
    STCN & 81.1 & 85.4 & 78.2 & 85.9 & 82.7 & 82.2 & 88.6 & 85.4 & 72.7 & 79.6 & 76.1 & 19.5\\
    AOT-L & 83.5 & 88.1 & 78.4 & 86.3 & 84.1 & 82.3 & 87.5 & 84.9 & 75.9 & 83.3 & 79.6 & 18.0 \\
    AOT-L\textsuperscript{\textdagger} & 84.0 & 88.8 & 78.4 & 86.7 & 84.5 & 82.4 & 88.4 & 85.4 & 77.3 & {85.1} & 81.2 & 12.1 \\
    DeAOT-L & 84.6 & 89.4 & 80.8 & 88.9 & 85.9 & 82.2 & 88.2 & 85.2 & 76.9 & 84.5 & 80.7 & 34.0\textsuperscript{\textdaggerdbl} \\
    DeAOT-L\textsuperscript{\textdagger} & 85.3 & 90.2 & 80.4 & 88.6 & \textbf{86.1} & 83.1 & 89.2 & 86.2 & 78.9 & 86.7 & \textbf{82.8} & 21.1\textsuperscript{\textdaggerdbl} \\
    XMem & 84.3 & 88.6 & 80.3 & 88.6 & 85.5 & 82.9 & 89.5 & 86.2 & 77.4 & 84.5 & 81.0 & 34.4\textsuperscript{\textdaggerdbl} \\
    ISVOS & 85.2 & 89.7 & 80.7 & 88.9 & \textbf{86.1} & 83.7 & 90.5 & \textbf{87.1} & 79.3 & 86.2 & \textbf{82.8} & - \\
    \midrule
    \textbf{DeVOS-B} & 84.5 & 89.5 & 79.4 & 87.4 & 85.2 & 83.4 & 88.8 & 86.1 & 77.2 & 84.7 & 81.0 & 36.7\textsuperscript{\textdaggerdbl} \\
    \textbf{DeVOS-L} & 85.2 & 90.1 & 80.7 & 89.0 & \textbf{86.3} & 84.2 & 91.2 & \textbf{87.7} & 79.4 & 86.4 & \textbf{82.9} & 24.7\textsuperscript{\textdaggerdbl} \\
    \midrule
    \textbf{DeVOS-B\textsuperscript{\textasteriskcentered}} & 84.7 & 89.7 & 79.4 & 87.8 & 85.4 & 83.5 & 89.3 & 86.4 & 77.4 & 84.9 & 81.2 & 36.7\textsuperscript{\textdaggerdbl} \\
    \textbf{DeVOS-L\textsuperscript{\textasteriskcentered}} & 85.4 & 90.3 & 80.8 & 89.3 & \textbf{86.6} & 84.4 & 91.8 & \textbf{88.1} & 79.4 & 86.6 & \textbf{83.0} & 24.7\textsuperscript{\textdaggerdbl} \\
    \bottomrule[1.5pt]
\end{tabular}
\end{table*}

To study performance capabilities and contributions impact, we introduce two variants of network architecture. Namely, DeVOS-B (Base) is a baseline implementation of the proposed method featuring consistency with previous approaches and considerable runtime speeds. Alternatively, DeVOS-L (Large) is a larger-scaled configuration for which we adopt more advanced building blocks and inject more complex architecture decisions.

\paragraph{Encoder \& Decoder} To achieve fairness in comparison and to keep consistency with previous works \cite{yang2021associating, yang2021decoupling, cheng2022xmem, wang2022look}, we equip our basic model DeVOS-B with ImageNet1K \cite{imagenet15russakovsky} pre trained ResNet50 \cite{He2015} image feature encoder. Meanwhile, with the aim of enhancing instance understanding logic, our bigger model DeVOS-L is equipped with ViT-B \cite{dosovitskiy2020vit} encoder pre trained on Segment Anything Dataset \cite{kirillov2023segany}. We assume that large-scale pre-training of transformer encoder on supervised instance segmentation is more suitable for video object segmentation as it allows the backbone to learn the notion of what objects actually are. We leave this fact, though, for further research. FPN \cite{fpn2017} decoder with Group Normalization \cite{lin2017feature} is used in both DeVOS-B and DeVOS-L.

\paragraph{Object masks} Following \cite{cheng2021stcn, wang2022look}, we adopt a lightweight ResNet18 \cite{He2015} network to encode one-hot object masks into the multi-scale embedding space. The number of input channels to mask encoder is set to 15, matching the maximal object number in benchmarks. To achieve homogeneous and simultaneous learning of segmentation mask representation while training, the input one-hot mask is zero-padded to have 15 channels, and the objects (i.e., channels) are then randomly shuffled.

\vspace{-10pt}

\paragraph{Flow representation} Optical flow field is used to capture the motion context between consecutive frames. For this, we employ GMA \cite{jiang2021learning} network due to its favorable performance and flexibility in adjusting run-time speed by tweaking the number of refinement updates. Even though the original paper suggests performing 12 updates, we find that four is enough to provide a strong displacement prior to matching. Notably, our model is designed to be independent of the actual flow estimator implementation. To construct a multi-scale motion representation from estimated optical flow, a lightweight ResNet18 \cite{He2015} is used.

\section{Experiments}

\begin{figure*}[ht!]
    \centering
  \includegraphics[width=0.95\linewidth]{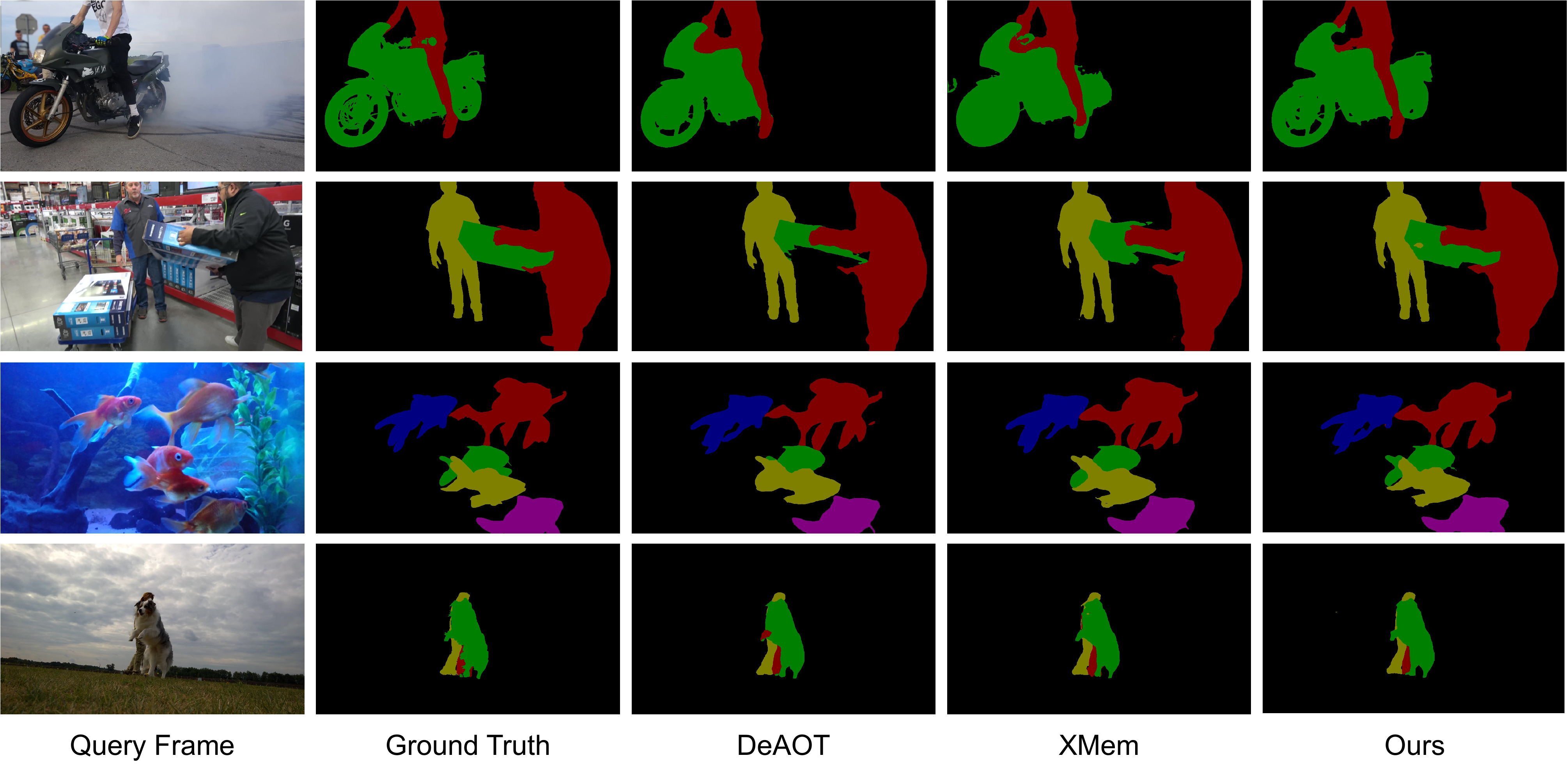}
  \caption{Qualitative comparison between DeVOS and some state-of-the-art VOS methods. Best viewed in zoom. We don't include ISVOS \cite{wang2022look} since there is no source code available. For all methods we used DAVIS2017 val sequences in 480p.}
  \label{fig:qualitative}
\end{figure*}

\subsection{Implementation details}

\paragraph{Training details}

Similarly to \cite{oh2019stm, cheng2021stcn, seong2021hierarchical, yang2021associating, yang2021decoupling, cheng2022xmem, wang2022look}, we split the training of DeVOS into two stages. During the first stage, we adopt pretraining on synthetic sequences derived from static image datasets \cite{wang2020end}. Consequently, we conduct main training on DAVIS 2017 \cite{Perazzi2016}, YouTubeVOS 2019 \cite{xu2018youtubevos1}, and optionally MOSE 2023 \cite{MOSE}. A more detailed description of training is provided in Supplementary.

\begin{table}[b!]
\caption{The quantitative evaluation on DAVIS 2016.}
\label{tab:davis2016}
\vspace{5pt}
\centering
  \begin{tabular}{l|ccc}
    \toprule[2pt]
    Methods & $\mathcal{J}$ & $\mathcal{F}$ & $\mathcal{J}\&\mathcal{F}$ \\
    \midrule[1.5pt]
    AOT-T & 86.1 & 87.4 & 86.8 \\
    DeAOT-T & 87.8 & 89.9 & 88.9 \\
    STCN & 90.8 & 92.5 & 91.6 \\
    XMem & 90.4 & 92.7 & 91.5 \\
    ISVOS & \textbf{91.5} & 93.7 & 92.6 \\
    Swin-B AOT-L & 90.7 & 93.3 & 92.0 \\
    Swin-B DeAOT-L & 91.1 & 94.7 & 92.9 \\
    \midrule
    \textbf{DeVOS-B} & 90.8 & 93.0 & 91.9 \\
    \textbf{DeVOS-L} & 91.0 & \textbf{95.8} & \textbf{93.5} \\
    \bottomrule[1.5pt]
\end{tabular}
\end{table}

\paragraph{Evaluation}
\label{sec:dataset}

In order to evaluate our models, we use traditional VOS metrics as proposed in \cite{Pont-Tuset_arXiv_2017}. We evaluate our method on DAVIS 2016 \& 2017 using the default 480p 24FPS videos, not benefiting from higher resolutions or test-time augmentations. The impact of multi-scale inference \cite{chandra2016fast} augmentation is studied in Supplementary. While evaluating our method on YouTube-VOS 2019 validation split, we exploit all intermediate frames of the videos to benefit from smooth motion implying more accurate optical flow. Even though we use 24 FPS sequences during evaluation, the 6FPS version is used during training and for metric computation.

\vspace{-10pt}

\paragraph{Inference}

Following \cite{cheng2021stcn, oh2019stm, cheng2022xmem, wang2022look}, we maintain feature memory by memorizing every fifth frame during inference. To keep consistent run-time speeds and stable memory consumption, the memory bank is implemented as a FIFO queue with a maximum size of 16. Meanwhile, the first frame is always kept in the memory \cite{wang2022look}. We don't use top-k filtering \cite{cheng2021mivos} or kernelized memory readouts \cite{Seong2020KernelizedMN} as we rely on short-term matching for smooth propagation and on QK-flow for temporal consistency.

\subsection{Comparison with State-of-the-art Methods}

\paragraph{Quantitative comparison} \Cref{tab:main} presents a comparison of DeVOS with other state-of-the-art methods on DAVIS 2017 validation, DAVIS 2017 test-dev, and Youtube-VOS 2019 validation. The quantitative comparison on DAVIS 2016 validation is listed in \Cref{tab:davis2016}.

We can see that without BL30K \cite{cheng2021mivos} for pretraining and MOSE \cite{MOSE} for main training, our ViT-B DeVOS-L achieves state-of-the-art performance scoring \textbf{87.7\%} $\mathcal{J}\&\mathcal{F}$ on DAVIS 2017 validation set, \textbf{82.9\%} $\mathcal{J}\&\mathcal{F}$ on DAVIS 2017 test set and \textbf{86.3\%} $\mathcal{J}\&\mathcal{F}$ on Youtube-VOS 2019 validation. The integration of the MOSE dataset further enhances our metrics, resulting in improved performance: \textbf{88.1\%} $\mathcal{J}\&\mathcal{F}$ \textbf{83.0\%} $\mathcal{J}\&\mathcal{F}$, \textbf{86.6\%} $\mathcal{J}\&\mathcal{F}$ on the DAVIS 2017 validation and test-dev, and on the Youtube-VOS 2019 validation.

The DeVOS-B model exhibits robust performance on the principal benchmark, scoring \textbf{86.1\%} $\mathcal{J}\&\mathcal{F}$ on DAVIS 2017 validation set, \textbf{81.0\%} $\mathcal{J}\&\mathcal{F}$ on DAVIS 2017 test set and \textbf{85.2\%} $\mathcal{J}\&\mathcal{F}$ on Youtube-VOS 2019 validation. Despite employing a less complex memory mechanism in contrast to \cite{cheng2022xmem}, omitting the direct injection of instance information as \cite{wang2022look}, and foregoing hierarchical propagation like \cite{yang2021decoupling}, our method achieves commendable outcomes in both accuracy and, notably, FPS. This highlights the efficacy of incorporating multi-scale matching and motion-guided attention, as it contributes to the enhancement of matching performance.


\begin{table*}[ht!]
    \centering
\caption{Ablation study. The experiments are based on DeVOS-B. MS: multi-scale matching. ADVA: adaptive deformable video attention. FP: flow priors to offset prediction in ADVA. QK: query-key flow enhancement. $N_k$: number of offsets per head / scale in deformable attention. Iters: number of flow refinement iterations of GMA. $\omega$: scale embedding. $\theta$: offset normalization. Note: in the final configuration QK-flow is used only in DeVOS-L.}
\label{tab:ablation}
\begin{subtable}{0.49\linewidth}\centering
\subcaption{Multi-scale matching}
\label{tab:ms}
  \begin{tabular}{cc|ccc|cc}
    \toprule[2pt]
    MS & ADVA & $D_{17V}$ & $D_{17T}$ & $Y_{19}$ & $\#\text{param}$ & FPS\\
    \midrule
    \ding{55} & \ding{55} & 84.7 & 79.2 & 83.7 & 35.4M & 32.4 \\
    \checkmark & \ding{55} & 85.2 & 80.5 & 84.4 & 38.1M & 12.9 \\
    \checkmark & \checkmark & 86.1 & 81.0 & 85.2 & 40.3M & 36.7 \\
    \bottomrule[1.5pt]
    \vspace{4pt}
\end{tabular}
\end{subtable}
\begin{subtable}{0.49\linewidth}\centering
\subcaption{Motion injection}
\label{tab:motion}
  \begin{tabular}{cc|ccc|cc}
    \toprule[2pt]
    FP & QK & $D_{17V}$ & $D_{17T}$ & $Y_{19}$ & $\#\text{param}$ & FPS\\
    \midrule
    \ding{55} & \ding{55} & 84.9 & 79.6 & 84.2 & 31.1M & 52.1 \\
    \checkmark & \ding{55} & 86.1 & 81.0 & 85.2 & 40.3M & 36.7 \\
    \checkmark & \checkmark & 86.5 & 81.2 & 85.3 & 40.4M & 29.4 \\
    \bottomrule[1.5pt]
    \vspace{4pt}
\end{tabular}
\end{subtable}
\begin{subtable}{0.33\linewidth}\centering
\subcaption{Number of offsets}
\label{tab:offsets}
  \begin{tabular}{c|cc|c}
    \toprule[2pt]
    $N_k$ & $D_{17V}$ & $Y_{19}$ & FPS\\
    \midrule
    2 & 82.5 & 81.8 & 37.1 \\
    4 & 86.1 & 85.2 & 36.7 \\
    6 & 86.3 & 84.9 & 36.1 \\
    8 & 86.0 & 84.6 & 35.5 \\
    \bottomrule[1.5pt]
\end{tabular}
\end{subtable}
\begin{subtable}{0.33\linewidth}\centering
\subcaption{Optical flow iters}
\label{tab:iters}
  \begin{tabular}{c|cc|c}
    \toprule[2pt]
    Iters & $D_{17V}$ & $Y_{19}$ & FPS\\
    \midrule
    0 & 81.3 & 81.0 & 45.5 \\
    1 & 85.5 & 84.9 & 41.4 \\
    4 & 86.1 & 85.2 & 36.7 \\
    12 & 86.3 & 85.3 & 24.8 \\
    \bottomrule[1.5pt]
\end{tabular}
\end{subtable}
\begin{subtable}{0.33\linewidth}\centering
\subcaption{Scale emb \& offset norm}
\label{tab:norm}
  \begin{tabular}{cc|cc}
    \toprule[2pt]
    $\omega$ & $\theta$ & $D_{17V}$ & $Y_{19}$ \\
    \midrule
    \ding{55} & \ding{55} & 84.9 & 84.1 \\
    \checkmark & \ding{55} & 85.0 & 84.3 \\
    \ding{55} & \checkmark & 85.8 & 85.0 \\
    \checkmark & \checkmark & 86.1 & 85.2 \\
    \bottomrule[1.5pt]
\end{tabular}
\end{subtable}
\end{table*}

\vspace{-10pt}

\paragraph{Qualitative comparison} \cref{fig:qualitative} displays the qualitative comparison of our method with recent state-of-the-art approaches. As shown by the gold-fish sequence, our approach demonstrates superior performance under complex shape and appearance changes. Additionally, mbike-trick sequence demonstrates that our design results in strong performance under rapid motion. These findings highlight the effectiveness of the proposed approach in handling various challenging conditions.

\subsection{Discussion}

\paragraph{Training with MOSE 2023} MOSE 2023 \cite{MOSE} (Co\textbf{M}plex video \textbf{O}bject \textbf{SE}gmentation) is a novel VOS benchmark featuring extreme scenarios of the video sequence which are not handled good enough by existing VOS methods. The main features of introduced videos include a large number of crowded and similar objects, heavy occlusions by similar-looking objects, extremely small-scale objects, and reference masks covering only a small region of the whole object.

With adopting MOSE 2023 as training data, performance on the classic benchmarks experiences only a small boost (\Cref{tab:main}), likely because they don't feature any similar extreme scenarios. However, DAVIS and Youtube-VOS focus on circumstances with a large number of object classes and classes unseen during training, along with a wide variety of challenging environments, while MOSE 2023 lacks such flexibility. Wrapping up, even minor improvements on classic benchmarks while training with MOSE 2023 indicate the high robustness and performance capacity of the proposed method. The quantitative comparison with other methods on MOSE 2023 validation is studied in Supplementary.

\vspace{-10pt}

\paragraph{Impact of multi-scale matching}
We argue that matching conducted solely on 1/16 of the input resolution does not convey enough spatial information and fine-grained details to perform instance discrimination effectively. This limitation becomes particularly crucial when dealing with overlapping objects that share a similar appearance. Conversely, the short-term branch can leverage smaller feature map resolutions, specifically 1/8. To validate these hypotheses, we remove multi-scale matching and evaluate the performance of the resulting architecture in \Cref{tab:ms}. Multi-scale matching increases $\mathcal{J}\&\mathcal{F}$ by \textbf{0.5\%} and ADVA matching further boosts the performance by \textbf{0.9\%} $\mathcal{J}\&\mathcal{F}$ while featuring $\times 3$ run-time speed boost on multi-scale. The importance of the number of sampled offsets per attention head and scale is studied in \Cref{tab:offsets}. Selecting $N_k=4$ provides optimal performance-efficiency tradeoff. Scale embedding and offset normalization drastically improve training stability thus lead to better final performance, which is reflected in \Cref{tab:norm}.

\vspace{-10pt}

\paragraph{Impact of optical flow guidance} 

We assert that to make matching emphasize semantic features and instance discrimination it is necessary to inject global motion understanding prior to the matching process. To accomplish this, we enhance offset prediction with optical flow. Additionally, we study the effect of QK-flow, which directly injects motion information into the query and value feature maps. We argue that this ensures strong cycle consistency. From \Cref{tab:motion}, we can see that removing QK-flow results in a reduction of \textbf{0.4\%} $\mathcal{J}\&\mathcal{F}$ on the DAVIS 2017 Validation set. Additionally, removing optical flow-based offset prediction results in a reduction in $\mathcal{J}\&\mathcal{F}$ by \textbf{1.2\%}.

\vspace{-11pt}

\paragraph{Impact of ViT backbone}

To further evaluate the impact of ViT-backbone, we train the same architecture but with Swin-B \cite{liu2021Swin} transformer used as the backbone. This results in decrease in $\mathcal{J}\&\mathcal{F}$ by \textbf{0.2\%}. We leave evaluation whether this small improvement comes from backbone architecture or SAM \cite{kirillov2023segany} pre-training for further research.

\vspace{-11pt}

\paragraph{Limitations}
One practical limitation is that the framework depends on a pre-trained optical flow estimator. We believe, though, that it is quite common that both optical flow and VOS are required simultaneously. Moreover, our approach works with different flow estimating architectures thus provides flexibility of actual choice (without need of retraining). Ablation on the number of optical flow iterations of GMA \cite{jiang2021learning} (\Cref{tab:iters}) shows that quality of optical flow is not crucial in the overall performance of our framework and thus any method performing good enough would work fine. 

\section{Conclusion}

This paper proposes DeVOS (Deformable VOS), an architecture that incorporates adaptive deformable video attention. DeVOS combines memory-based matching with motion-guided propagation, resulting in robust matching under challenging appearance changes and strong temporal consistency. DeVOS achieves state-of-the-art performance while maintaining top-rank FPS.  

{\small
\printbibliography

@String(IJCV = {Int. J. Comput. Vis.})

@String(CVPR= {IEEE Conf. Comput. Vis. Pattern Recog.})

@String(ICCV= {Int. Conf. Comput. Vis.})

@String(ECCV= {Eur. Conf. Comput. Vis.})

@String(ICLR = {Int. Conf. Learn. Represent.})

@String(IJCV  = {IJCV})

@String(CVPR  = {CVPR})

@String(ICCV  = {ICCV})

@String(ECCV  = {ECCV})

@String(ICLR  = {ICLR})

@article{horn1981determining,
  title={Determining optical flow},
  author={Horn, Berthold K and Schunck, Brian G},
  journal={Artificial intelligence},
  volume={17},
  number={1-3},
  pages={185--203},
  year={1981},
  publisher={Elsevier}
}

@inproceedings{black1993framework,
  title={A framework for the robust estimation of optical flow},
  author={Black, Michael J and Anandan, P},
  booktitle={Proceedings of the 4th International Conference on Computer Vision},
  pages={231--236},
  year={1993},
  organization={IEEE}
}

@article{bruhn2005lucas,
  title={Lucas/kanade meets horn/schunck: Combining local and global optic flow methods},
  author={Bruhn, Andr{\'e} and Weickert, Joachim and Schn{\"o}rr, Christoph},
  journal={International journal of computer vision},
  volume={61},
  number={3},
  pages={211--231},
  year={2005},
  publisher={Springer}
}

@article{sun2014quantitative,
  title={A quantitative analysis of current practices in optical flow estimation and the principles behind them},
  author={Sun, Deqing and Roth, Stefan and Black, Michael J},
  journal={International Journal of Computer Vision},
  volume={106},
  number={2},
  pages={115--137},
  year={2014},
  publisher={Springer}
}

@inproceedings{teed2020raft,
  title={RAFT: Recurrent all-pairs field transforms for optical flow},
  author={Teed, Zachary and Deng, Jia},
  booktitle={European Conference on Computer Vision},
  pages={402--419},
  year={2020},
  organization={Springer}
}

@article{jiang2021learning,
  title={Learning to estimate hidden motions with global motion aggregation},
  author={Jiang, Saining and Campbell, Dylan and Lu, Yi and Li, Hongdong and Hartley, Richard},
  journal={arXiv preprint arXiv:2104.02409},
  year={2021}
}

@article{bai2022deep,
  title={Deep Equilibrium Optical Flow Estimation},
  author={Bai, Shaojie and Geng, Zhaoyang and Savani, Yash and Kolter, J Zico},
  journal={arXiv preprint arXiv:2204.08442},
  year={2022}
}

@article{huang2022flowformer,
  title={FlowFormer: A Transformer Architecture for Optical Flow},
  author={Huang, Ziqi and Shi, Xuesong and Zhang, Chenxu and Wang, Qiyang and Cheung, Kin Chung and Qin, Hong and Dai, Jifeng and Li, Hao},
  journal={arXiv preprint arXiv:2203.16194},
  year={2022}
}

@article{jaegle2021perceiver,
  title={Perceiver IO: A general architecture for structured inputs \& outputs},
  author={Jaegle, Andrew and Borgeaud, Sébastien and Alayrac, Jean-Baptiste and Doersch, Carl and Ionescu, Catalin and Ding, Dov and Koppula, Skanda and Zoran, Dan and Brock, Andrew and Shelhamer, Evan and others},
  journal={arXiv preprint arXiv:2107.14795},
  year={2021}
}

@inproceedings{vaswani2017attention,
  title={Attention is all you need},
  author={Vaswani, Ashish and Shazeer, Noam and Parmar, Niki and Uszkoreit, Jakob and Jones, Llion and Gomez, Aidan N and Kaiser, Lukasz and Polosukhin, Illia},
  booktitle={Advances in neural information processing systems},
  pages={5998--6008},
  year={2017}
}

@article{xie2021rmnet,
  title={RMNet: Equivalently Removing Residual Connection from Networks},
  author={Xie, Haoxiang and Wang, Wenhai and Li, Xiang and Xie, Lingxi and Zhang, Ya and Tian, Qi},
  journal={arXiv preprint arXiv:2111.00687},
  year={2021}
}

@article{yang2021associating,
  title={Associating Objects with Transformers for Video Object Segmentation},
  author={Yang, Zongxin and Wei, Yunchao and Yang, Yi},
  journal={arXiv preprint arXiv:2106.02638},
  year={2021}
}

@article{yang2021decoupling,
  title={Associating Objects with Transformers for Video Object Segmentation},
  author={Yang, Zongxin and Yang, Yi},
  journal={arXiv preprint arXiv:2210.09782},
  year={2022}
}

@inproceedings{cheng2022xmem,
  title={{XMem}: Long-Term Video Object Segmentation with an Atkinson-Shiffrin Memory Model},
  author={Cheng, Ho Kei and Alexander G. Schwing},
  booktitle={ECCV},
  year={2022}
}

@misc{wang2022look,
      title={Look Before You Match: Instance Understanding Matters in Video Object Segmentation}, 
      author={Junke Wang and Dongdong Chen and Zuxuan Wu and Chong Luo and Chuanxin Tang and Xiyang Dai and Yucheng Zhao and Yujia Xie and Lu Yuan and Yu-Gang Jiang},
      year={2022},
      eprint={2212.06826},
      archivePrefix={arXiv},
      primaryClass={cs.CV}
}

@misc{lin2017feature,
      title={Feature Pyramid Networks for Object Detection}, 
      author={Tsung-Yi Lin and Piotr Dollár and Ross Girshick and Kaiming He and Bharath Hariharan and Serge Belongie},
      year={2017},
      eprint={1612.03144},
      archivePrefix={arXiv},
      primaryClass={cs.CV}
}

@inproceedings{liu2021Swin,
  title={Swin Transformer: Hierarchical Vision Transformer using Shifted Windows},
  author={Liu, Ze and Lin, Yutong and Cao, Yue and Hu, Han and Wei, Yixuan and Zhang, Zheng and Lin, Stephen and Guo, Baining},
  booktitle={Proceedings of the IEEE/CVF International Conference on Computer Vision (ICCV)},
  year={2021}
}

@article{Pont-Tuset_arXiv_2017,
  author = {Jordi Pont-Tuset and Federico Perazzi and Sergi Caelles and Pablo Arbel\'aez and Alexander Sorkine-Hornung and Luc {Van Gool}},
  title = {The 2017 DAVIS Challenge on Video Object Segmentation},
  journal = {arXiv:1704.00675},
  year = {2017}
}

@misc{xu2018youtubevos1,
      title={YouTube-VOS: A Large-Scale Video Object Segmentation Benchmark}, 
      author={Ning Xu and Linjie Yang and Yuchen Fan and Dingcheng Yue and Yuchen Liang and Jianchao Yang and Thomas Huang},
      year={2018},
      eprint={1809.03327},
      archivePrefix={arXiv},
      primaryClass={cs.CV}
}

@article{MOSE,
  title={MOSE: A New Dataset for Video Object Segmentation in Complex Scenes},
  author={Ding, Henghui and Liu, Chang and He, Shuting and Jiang, Xudong and Torr, Philip HS and Bai, Song},
  journal={arXiv preprint arXiv:2302.01872},
  year={2023}
}

@inproceedings{Perazzi2016,
  author = {F. Perazzi and J. Pont-Tuset and B. McWilliams and L. {Van Gool} and M. Gross and A. Sorkine-Hornung},
  title = {A Benchmark Dataset and Evaluation Methodology for Video Object Segmentation},
  booktitle = {Computer Vision and Pattern Recognition},
  year = {2016}
}

@misc{chandra2016fast,
      title={Fast, Exact and Multi-Scale Inference for Semantic Image Segmentation with Deep Gaussian CRFs}, 
      author={Siddhartha Chandra and Iasonas Kokkinos},
      year={2016},
      eprint={1603.08358},
      archivePrefix={arXiv},
      primaryClass={cs.CV}
}

@article{He2015,
	author = {Kaiming He and Xiangyu Zhang and Shaoqing Ren and Jian Sun},
	title = {Deep Residual Learning for Image Recognition},
	journal = {arXiv preprint arXiv:1512.03385},
	year = {2015}
}

@article{imagenet15russakovsky,
    Author = {Olga Russakovsky and Jia Deng and Hao Su and Jonathan Krause and Sanjeev Satheesh and Sean Ma and Zhiheng Huang and Andrej Karpathy and Aditya Khosla and Michael Bernstein and Alexander C. Berg and Li Fei-Fei},
    Title = { {ImageNet Large Scale Visual Recognition Challenge} },
    Year = {2015},
    journal   = {International Journal of Computer Vision (IJCV)},
    doi = {10.1007/s11263-015-0816-y},
    volume={115},
    number={3},
    pages={211-252}
}

@article{dosovitskiy2020vit,
  title={An Image is Worth 16x16 Words: Transformers for Image Recognition at Scale},
  author={Dosovitskiy, Alexey and Beyer, Lucas and Kolesnikov, Alexander and Weissenborn, Dirk and Zhai, Xiaohua and Unterthiner, Thomas and  Dehghani, Mostafa and Minderer, Matthias and Heigold, Georg and Gelly, Sylvain and Uszkoreit, Jakob and Houlsby, Neil},
  journal={ICLR},
  year={2021}
}

@article{kirillov2023segany,
  title={Segment Anything},
  author={Kirillov, Alexander and Mintun, Eric and Ravi, Nikhila and Mao, Hanzi and Rolland, Chloe and Gustafson, Laura and Xiao, Tete and Whitehead, Spencer and Berg, Alexander C. and Lo, Wan-Yen and Doll{\'a}r, Piotr and Girshick, Ross},
  journal={arXiv:2304.02643},
  year={2023}
}

@inproceedings{cheng2021stcn,
  title={Rethinking Space-Time Networks with Improved Memory Coverage for Efficient Video Object Segmentation},
  author={Cheng, Ho Kei and Tai, Yu-Wing and Tang, Chi-Keung},
  booktitle={NeurIPS},
  year={2021}
}

@inproceedings{cheng2021mivos,
  title={Modular Interactive Video Object Segmentation: Interaction-to-Mask, Propagation and Difference-Aware Fusion},
  author={Cheng, Ho Kei and Tai, Yu-Wing and Tang, Chi-Keung},
  booktitle={CVPR},
  year={2021}
}

@article{zhu2020deformable,
  title={Deformable DETR: Deformable Transformers for End-to-End Object Detection},
  author={Zhu, Xizhou and Su, Weijie and Lu, Lewei and Li, Bin and Wang, Xiaogang and Dai, Jifeng},
  journal={arXiv preprint arXiv:2010.04159},
  year={2020}
}

@InProceedings{Xia_2022_CVPR,
    author    = {Xia, Zhuofan and Pan, Xuran and Song, Shiji and Li, Li Erran and Huang, Gao},
    title     = {Vision Transformer With Deformable Attention},
    booktitle = {Proceedings of the IEEE/CVF Conference on Computer Vision and Pattern Recognition (CVPR)},
    month     = {June},
    year      = {2022},
    pages     = {4794-4803}
}

@INPROCEEDINGS{oh2019stm,

  author={Oh, Seoung Wug and Lee, Joon-Young and Xu, Ning and Kim, Seon Joo},

  booktitle={2019 IEEE/CVF International Conference on Computer Vision (ICCV)}, 

  title={Video Object Segmentation Using Space-Time Memory Networks}, 

  year={2019},

  volume={},

  number={},

  pages={9225-9234},

  doi={10.1109/ICCV.2019.00932}}

@inproceedings{seong2021hierarchical,
  title={Hierarchical Memory Matching Network for Video Object Segmentation},
  author={Seong, Hongje and Oh, Seoung Wug and Lee, Joon-Young and Lee, Seongwon and Lee, Suhyeon and Kim, Euntai},
  booktitle={Proceedings of the IEEE/CVF International Conference on Computer Vision},
  year={2021}
}

@inproceedings{yang2020CFBI,
  title={Collaborative video object segmentation by foreground-background integration},
  author={Yang, Zongxin and Wei, Yunchao and Yang, Yi},
  booktitle={European Conference on Computer Vision},
  pages={332--348},
  year={2020},
  organization={Springer}
}

@article{yang2020CFBIP,
  author={Yang, Zongxin and Wei, Yunchao and Yang, Yi},
  journal={IEEE Transactions on Pattern Analysis and Machine Intelligence}, 
  title={Collaborative Video Object Segmentation by Multi-Scale Foreground-Background Integration}, 
  year={2021},
  volume={},
  number={},
  pages={1-1},
  doi={10.1109/TPAMI.2021.3081597}
}

@inproceedings{NEURIPS2020_liangVOS,
 author = {Liang, Yongqing and Li, Xin and Jafari, Navid and Chen, Jim},
 booktitle = {Advances in Neural Information Processing Systems},
 editor = {H. Larochelle and M. Ranzato and R. Hadsell and M. F. Balcan and H. Lin},
 pages = {3430--3441},
 publisher = {Curran Associates, Inc.},
 title = {Video Object Segmentation with Adaptive Feature Bank and Uncertain-Region Refinement},
 url = {https://proceedings.neurips.cc/paper/2020/file/234833147b97bb6aed53a8f4f1c7a7d8-Paper.pdf},
 volume = {33},
 year = {2020}
}

@inproceedings{wang2020end,
  title={End-to-End Video Instance Segmentation with Transformers},
  author={Wang, Yuqing and Xu, Zhaoliang and Wang, Xinlong and Shen, Chunhua and Cheng, Baoshan and Shen, Hao and Xia, Huaxia},
  booktitle =  {Proc. IEEE Conf. Computer Vision and Pattern Recognition (CVPR)},
  year={2021}
}

@article{Seong2020KernelizedMN,
  title={Kernelized Memory Network for Video Object Segmentation},
  author={Hongje Seong and Junhyuk Hyun and Euntai Kim},
  journal={ArXiv},
  year={2020},
  volume={abs/2007.08270}
}

@INPROCEEDINGS{fpn2017,
  author={Lin, Tsung-Yi and Dollár, Piotr and Girshick, Ross and He, Kaiming and Hariharan, Bharath and Belongie, Serge},
  booktitle={2017 IEEE Conference on Computer Vision and Pattern Recognition (CVPR)}, 
  title={Feature Pyramid Networks for Object Detection}, 
  year={2017},
  volume={},
  number={},
  pages={936-944},
  doi={10.1109/CVPR.2017.106}
}

@InProceedings{Fedynyak_2023_CVPR,
    author    = {Fedynyak, Volodymyr and Romanus, Yaroslav and Dobosevych, Oles and Babin, Igor and Riazantsev, Roman},
    title     = {Global Motion Understanding in Large-Scale Video Object Segmentation},
    booktitle = {Proceedings of the IEEE/CVF Conference on Computer Vision and Pattern Recognition (CVPR) Workshops},
    month     = {June},
    year      = {2023},
    pages     = {3152-3161}
}
}

\end{document}